\documentclass[letterpaper]{article}

\usepackage{amsmath}
\usepackage{natbib,alifeconf}  %% The order is important
\usepackage{url,hyperref,cleveref}
\usepackage{booktabs}
\usepackage{color}

\newcommand\blfootnote[1]{%
  \begingroup
  \renewcommand\thefootnote{}\footnote{#1}%
  \addtocounter{footnote}{-1}%
  \endgroup
}

\title{Learning Developmental Scaffoldings to Guide Self-Organisation}
\author{
    Milton L. Montero $^{1}$,
    Elias Najarro $^1$,
    Jakob Schauser$^2$ \and
    Sebastian Risi$^{1,3}$
    \mbox{}\\
    $^1$IT University of Copenhagen, Denmark \\
    $^2$University of Copenhagen, Denmark \\
    $^3$Sakana AI, Japan \\
    \{mlle,enaj,sebr\}@itu.dk, jakob.schauser@nbi.ku.dk
} % email of corresponding author

\begin{document}
\maketitle

\begin{abstract}
From subcellular structures to entire organisms, many natural systems generate complex organisation through self-organisation: local interactions that collectively give rise to global structure without any blueprint of the outcome. Yet a significant portion of the information driving such processes is not produced by self-organisation itself, instead, it is often offloaded to initial conditions of the system, analogous to a memory-compute trade-off in computational systems. Biological development is a prime example, where maternal pre-patterns encode positional and symmetry-breaking information that scaffolds the self-organising process. In this work, we study this offloading phenomenon by introducing a model that jointly learns both the self-organisation rules and the pre-patterns, allowing their interplay to be varied and measured under controlled conditions: a Neural Cellular Automaton (NCA) paired with a learned coordinate-based pattern generator (SIREN), both trained simultaneously to generate a set of patterns. We provide information-theoretic analyses of how information is distributed between pre-patterns and the self-organising process, and show that jointly learning both components yields improvements in robustness, encoding capacity, and symmetry breaking over purely self-organising alternatives. Our analysis further suggests that effective pre-patterns do not simply approximate their targets; rather, they bias the developmental dynamics in ways that facilitate convergence, pointing to a non-trivial relationship between the structure of initial conditions and the dynamics of self-organisation.
\end{abstract}

% Choose one of: Full Paper, Summaries, or Late Breaking Abstracts 
% Submission type: \textbf{Full Paper}\\

% If sharing code / data, anonymize your repository and paste the link here.
% Example of anonymizing sevice for github: https://anonymous.4open.science/
% delete this line if not needed
Data/Code available at: \url{https://github.com/miltonllera/siren-growth}. 
\blfootnote{\textcopyright  2026 [AUTHORS' NAMES]. Published under a Creative Commons Attribution 4.0 International (CC BY 4.0) license.}

\section{Introduction}

Self-organisation is widely regarded as one of the fundamental concepts in the study of complex systems \citep{weber_origins_1998}. It describes how global structure and function can emerge from local interactions between components, without the need for a central controller or explicit blueprint. A fundamental, yet often under explored, question in the study of self-organising systems is: \emph{where does the information that drives this process come from?} In many natural systems, that information is not generated entirely by the self-organising process itself, instead, some of it is offloaded to the system's initial conditions.

\begin{figure*}[!t]
    \centering
    \includegraphics[width=\linewidth]{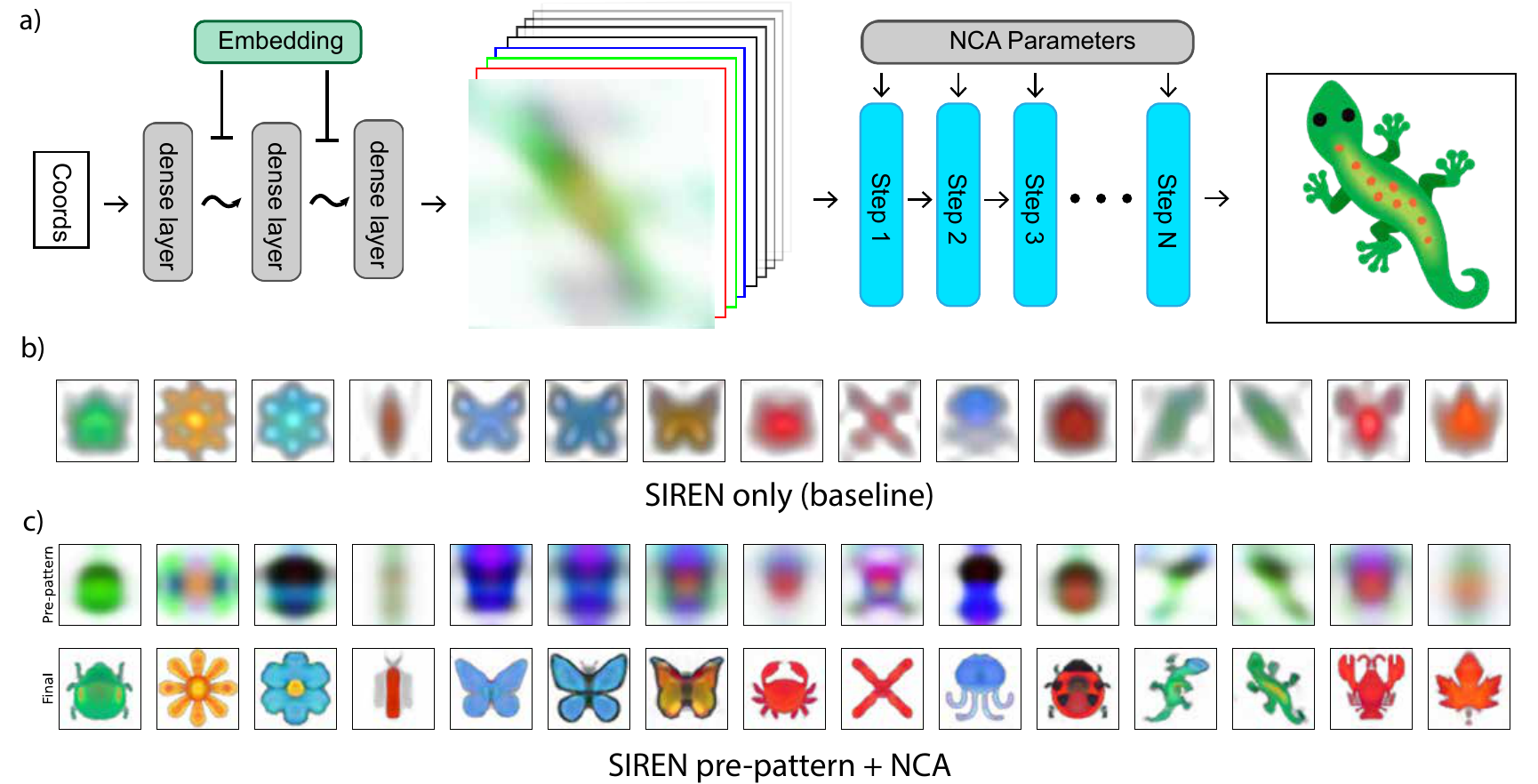}
    \vspace{-5mm}
    \caption{\textbf{Model diagram and results examples.} \textbf{a)} A pre-pattern function, defined using an implicit representation, transforms coordinates into initial values of the visible channels of the NCA. Development then unfolds using it as initial conditions towards the corresponding target. \textbf{b)} Optimal patterns for a one layer SIREN trained to directly predict the target showing that they are not capable of directly reconstructing the pattern. \textbf{c)} Example pre-patterns and final reconstructions when combining SIRENs (first row) and NCA (second row). In this case, information is shared between the two components, causing the SIREN output to less closely mimic the target, with the generated patterns instead becoming coarse representations of it.}
    \label{fig:model-def}
    \vspace{-0.5cm}
\end{figure*}

Biological development is a classical example of this phenomenon. Cell division, differentiation, and morphogenesis are all governed by local cell-cell interactions, yet this process does not operate in an informational vacuum. In many species, a substantial portion of the organisational information guiding development is instantiated via the initial conditions of the developmental substrate in the form of spatially structured pre-patterns: morphogenetic gradients, maternally provided mRNAs, and localised protein concentrations created by the parent organism prior to or during early development, which provide initial organisational information in the form of maternal effect genes \citep{gilbert_developmental_nodate}, body-axes specification \citep{lander_morpheus_2007}, and tissue level processes such as gastrulation \citep{inomata_scaling_2017} and limb formation \citep{raspopovic_digit_2014,meinhardt_boundary_1983} (see \citep{meinhardt_models_2008} for a broader overview). 

In all these cases morphogenetic gradients provide organisational information by enabling symmetry-breaking and providing positional information \citep{wolpert_positional_2016,green_positional_2015}. From the perspective of information theory, and in analogy to the memory-compute trade-off in computer science, this offloading reduces the informational burden on the self-organising process by transferring part of it into the system's starting state. This enables a key property: different pre-patterns can bias the subsequent unfolding of a single, shared developmental program \citep{mosby_morphogens_2024}, unlocking greater levels of compositionality and modularity in developmental space.

% These maternal-effect mechanisms break symmetry and provide positional information to the embryo \citep{grossniklaus_three_1994}, effectively pre-computing a portion of the organisational information that would otherwise have to be generated by the self-organising process from scratch. From the perspective of information theory, and in analogy to the memory-compute trade-off in computer science, this offloading reduces the informational burden on the self-organising process by transferring part of it into the system's starting state.

% Importantly, this transfer of information to initial conditions is not unique to early embryogenesis: it recurs across spatial and temporal scales, from the establishment of body axes to tissue-level processes such as gastrulation \citep{inomata_scaling_2017} and limb formation \citep{raspopovic_digit_2014, meinhardt_boundary_1983} (see \citep{meinhardt_models_2008} for a broader overview). In each case, a pre-pattern generated at a higher level of organisation scaffolds the self-organising dynamics that follow.  This compositionality, whereby variation in initial conditions propagates into diverse developmental outcomes through a shared process, is widely regarded as a source of modularity and a candidate route to evolvability in biological systems.

Despite its importance, studying the offloading of information to initial conditions in a controlled and quantitative manner is challenging. Molecular and developmental biology approaches characterise the phenomenon by identifying the specific genes, proteins, and gradients involved; information-theoretic and theoretical biology approaches model it at an abstract level. Both have yielded important insights \citep{arias_molecular_2013}, but neither allows one to systematically vary the balance between pre-patterned information and self-organisation and measure the downstream consequences. Doing so requires a model in which both components (the pre-pattern and the self-organisation rules) can be learned jointly under a common objective, enabling their relative contributions to be manipulated and assessed experimentally.

In this work, we study this offloading phenomenon using Neural Cellular Automata (NCA) \citep{mordvintsev_growing_2020} as the self-organising component: a class of models where all structural and positional information must be inferred through the dynamics of self-organisation itself, making them a natural baseline for studying the effects of pre-patterning. We pair the NCA with a learned coordinate-based pattern generator (SIREN) \citep{sitzmann_implicit_2020}, both trained simultaneously on a set of target patterns. The SIREN generates spatially structured initial conditions (the scaffold) from which the NCA self-organises toward each target, directly analogous to the biological mechanism described above. We analyse how information is distributed between the pre-pattern and the self-organising process using information-theoretic measures, and evaluate the system's robustness, encoding capacity, and symmetry breaking relative to a purely self-organising alternative. Our results show that offloading information to initial conditions improves all three properties. Moreover, effective pre-patterns do not simply approximate their targets; rather, they bias the developmental dynamics in ways that facilitate convergence, pointing to a non-trivial relationship between the structure of initial conditions and the dynamics of self-organisation.

\section{Methods}

\paragraph{Neural Cellular Automata} Neural Cellular Automata (NCA) \citep{mordvintsev_growing_2020} provide a natural starting point for a model of emergent behaviour. NCA extend classical cellular automata with learnable update rules, enabling the study of self-organisation in a framework amenable to gradient-based optimisation. They have been used to study developmental dynamics \citep{mordvintsev_growing_2020}, homeotic transformations \citep{hintze_gene-regulated_nodate}, distributed processing of sensory information \citep{kvalsund_sensor_2026}, reasoning \citep{guichard_arc-nca_2025}, distributed computation \citep{bena_path_2025}, and the evolvability of developmental programs \citep{montero_meta-learning_2024}. Crucially, in their standard formulation, NCA are \emph{purely} self-organising: all structural and positional information must be inferred by cells as development unfolds, with no pre-specified spatial structure in the initial state. This makes them a clean baseline against which the effects of offloading information to initial conditions can be assessed.

% \subsubsection{Model}
\subsection{Learning Developmental Scaffoldings}

Our system (seen in Figure~\ref{fig:model-def}) has two main components: a pre-pattern generator and a self-organising system which uses the pre-pattern as initial conditions which drive its update process. The self-organising component is implemented using an NCA with an isotropic perception kernel \cite{mordvintsev_growing_2020}. While usually hard to fit when used with standard NCAs due to the lack of symmetry breaking mechanisms, the use of pre-patterns successfully overcomes this issue while reducing the number of parameters in the NCA.

% Typically, using an isotropic kernel presents some complications because cells cannot easily break symmetry without positional cues (e.g. boundary conditions or noise random fluctuations). As we will see a spatial pre-pattern solves this issue quite effectively.

For the pre-pattern generator we use an implicit representation with periodic activation functions, or SIREN \citep{sitzmann_implicit_2020}. It generates smooth patterns from coordinate-based inputs and is compatible with gradient-based optimization, unlike approaches such as Compositional Pattern-Producing Networks \citep{stanley_compositional_2007}.

% For the pre-pattern generator any function capable of generating spatially coherent patterns is, in theory, valid. However, in order to match the type of continuous patterns we observe in biological systems, we want said patterns to vary smoothly in space.

% One popular option would be to use Compositional Pattern Producing Networks (CPPNs, \cite{stanley_compositional_2007}). CPPNs are capable of creating smooth spatially arranged patterns by design which they achieve by taking spatial coordinates as inputs. However, co-optimizing them along the NCA can prove challenging because the former is traditionally trained using NEAT while the latter is more amenable to back-propagation.

% Thus, we instead propose using an approximate approach using implicit representations with periodic activation functions, or SIREN \citep{sitzmann_implicit_2020}. Briefly, a SIREN is a multi-layer-perceptron which takes coordinates as inputs like a CPPN (hence the implicit representation moniker), and uses sinusoidal activation functions in the hidden layers. Importantly, it is capable of approximating the smooth patterns generated by CPPNs while being relatively easy to train thanks to the use of sine activation functions.

In more formal terms, a SIREN computes the following:
\begin{equation}
   (x, y) \mapsto f_{\theta}(x, y) 
\end{equation}
where $\theta$ are the parameters of the function. Standard formulation is to use an $L$-layer MLP with sinusoidal activation functions in the intermediate layers:
\begin{equation}
    h_{l+1} = \sin(\mathbf{W}_{l}\mathbf{h_l} + \mathbf{b}_l)
\end{equation}
where $h_0 = [x; y]$. Typically the output layer uses no activation function, however in our case we will use a sigmoidal activation function to restrict the four output channels (RGBA) to the $(0, 1)$ range. With $W_{L} \in \mathcal{M}_{[4, .]}$ we have:
\begin{equation}
    \text{RGBA}_{t=0} = \sigma(\mathbf{W}_{L}\mathbf{h}_L)
\end{equation}
Notice that the above formulation can only generate one particular pre-pattern, which is likely inappropriate for generating many different examples with one unified set of dynamics. One option would be to use goal-directed NCA \citep{sudhakaran_goal-guided_2022}, however here we opt instead for manipulating the starting conditions, which has also been shown to elicit different patterns from a single shared NCA rule \citep{Sinapayen2023May}. We can do this using a modulated SIREN formulation \citep{mehta_modulated_2021}. Here an embedding model (either pre-trained or trained simultaneously) generates per-target embeddings which can then be used to generate scaling factors for each hidden unit:
\begin{equation}
    \mathbf{m}_l = \text{ReLU}(\mathbf{M}_l \mathbf{e})
\end{equation}
where the $\mathbf{M}_l$'s have the same number of rows as the corresponding $\mathbf{W}_l$'s. Notice that the ReLU activation means that the output is in the range $(0, \infty)$. Thus we can scale the hidden activations of the SIREN model using the modulation vectors. With a slight abuse of notation we now have:
\begin{equation}
    \mathbf{h}_l = \mathbf{m}_l \cdot \mathbf{h}_l
\end{equation}
which enables the system to generate different patterns for different targets (see Figure~\ref{fig:model-def} for examples).

The initialisation is completed by taking the four initialised visible channels and concatenating $c$ zero-valued hidden channels:
\begin{equation}
    s_0 = \text{cat}(\sigma(h_L), \mathbf{0}_c)
\end{equation}
where we use the sigmoid activation function to normalize the visible channels to the valid RGBA range of $[0, 1]$.

The rest of the model proceeds as in the standard NCA formulation, alternating perception and update steps until we obtain $s_T$. The update function uses the usual MLP update with $128$ hidden units. The embedding model is a simple convolutional encoder which we train simultaneously.

\begin{figure*}[t!]
    \centering
    \includegraphics[width=\linewidth]{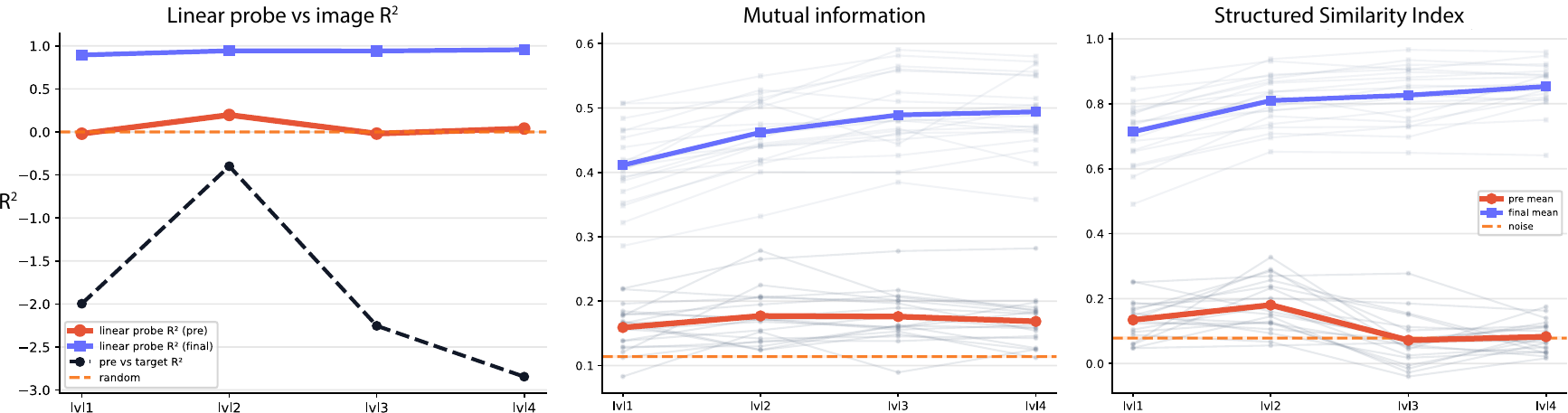}
    \caption{\textbf{Information trade-off}. We measure how information is balanced between the pre-pattern and the final reconstruction using three different measures. \textbf{\emph{Left:}} We apply a pixel-wise linear probe and compute the $R^2$ of the regression between the both pre and final patterns and the target. The plots show that at the start the pre-pattern is below chance, while at the end of development the patterns are very close to the target. \textbf{\emph{Centre:}} Similar to the previous plot but using Normalised Mutual Information (NMI) instead of a linear probe. The plot shows that the information content is not 0, but also doesn't increase with the complexity of the pre-pattern generating function. \textbf{\emph{Right:}} The same analysis as before performed using Structural Similarity Index Measure (SSIM), showing similar results.}
    \label{fig:analysis-info-trade}
    \vspace{-0.5cm}
\end{figure*}

\subsection{Training}

We train the system end-to-end using Stochastic Gradient Descent and a learning rate of $10^{-4}$ with Nesterov momentum \cite{bottou-91c}. The optimization criterion is Mean-Squared Error (MSE) between target and NCA output. Importantly, we do not use a trajectory pool to ensure stability even though the NCA is isotropic. In all cases training ran for $100{,}000$ update steps.

\subsection{Dataset} We use a dataset of 20 hand-crafted emojis as proxies for morphological structures. Importantly, they are designed to possess a variety of symmetries (radial and axial) and structural similarities. For example, some of them represent different types of flowers with similar symmetrical petals around the centre but varying in their colour and shape. 

\section{Experiments}

We evaluate our model across three sets of experiments designed to study how offloading information to initial conditions affects the properties of self-organising systems: the distribution of information between pre-patterns and development, robustness to cellular noise, and encoding capacity. Symmetry breaking, the third property highlighted in the introduction, is established implicitly throughout: our isotropic NCA cannot break symmetry without positional cues, and the pre-pattern supplies them (see Methods). For the latter two we compare against a goal-directed NCA (GoalNCA) \citep{sudhakaran_goal-guided_2022}, a purely self-organising alternative that generates multiple targets without the benefit of a pre-pattern. We additionally extend the model to growing systems, where the interplay between pre-patterning and self-organisation takes on a more explicitly biological character. 

Figure~\ref{fig:model-def}c illustrates how a fitted model behaves, plotting both the corresponding pre-patterns and final reconstructions for example targets. Crucially, the pre-patterns converge to coarse representations of the input, which is a consequence of only optimizing with respect to the final step of the NCA. Much detail is missing, but the structures tend to capture the different symmetries such as in the case of the blossom or X, and structure repetition as in the butterflies. In the case of this particular subclass we also see a strong resemblance between their respective pre-pattern. Taken together, these two facts point to models being able to reuse information in their initial conditions where needed: wings on both sides of the vertical axis of symmetry receive the same information since they essentially must build the same component and the same principle applies across species.

\begin{figure*}[t!]
    \centering
    \includegraphics[width=\linewidth]{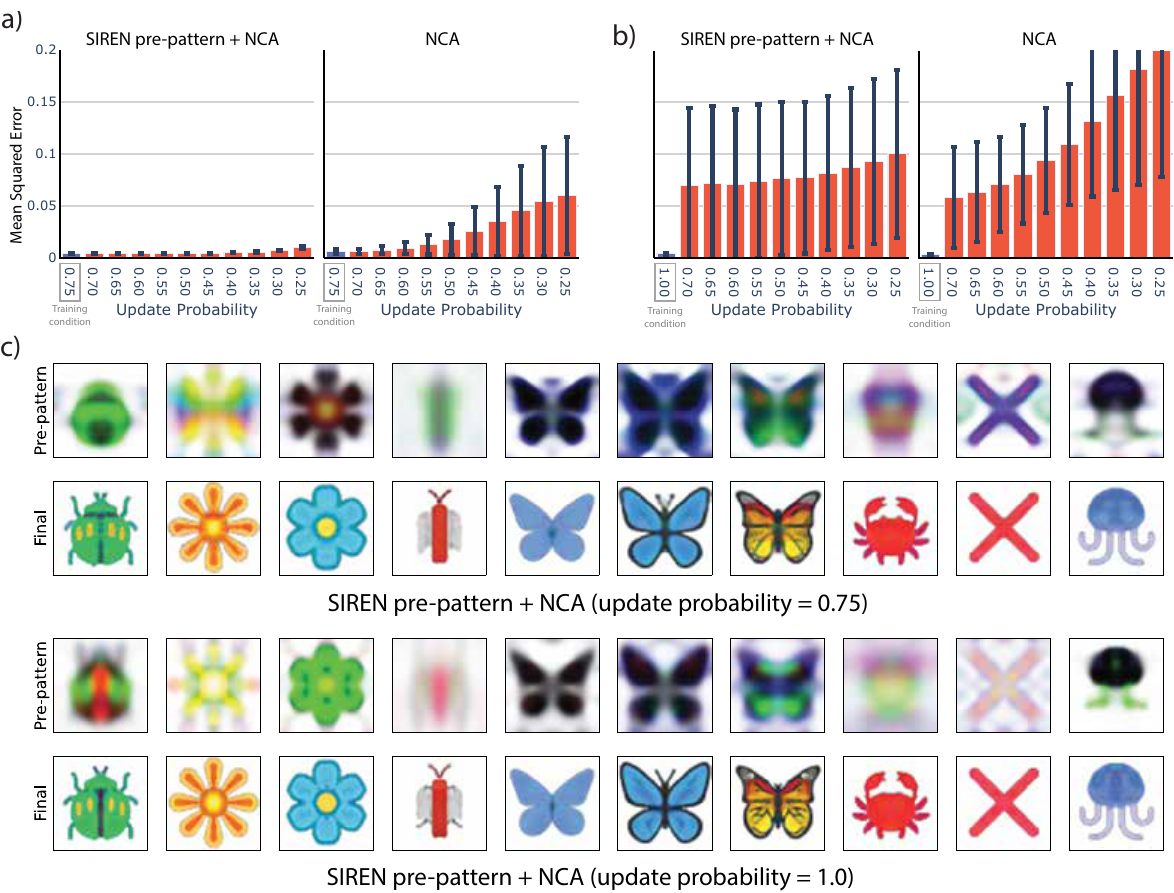}
    \caption{\textbf{Robustness to noise}. We compare the NCA with SIREN pre-patterns to a control model (GoalNCA, see main text), which can also generate multiple targets but uses self-organisation only, across multiple levels of reliability after training. \textbf{a)} The results for the NCA with pre-patterns on the left, on the right the control. As we decrease the reliability of the cells, the pre-patterned NCA suffers very little in terms of performance while GoalNCA quickly diverges. \textbf{b)} When both models are trained with a deterministic update (update probability $1.0$), they both fail to generalise to increased unreliability. \textbf{c)} Examples of pre-patterns and reconstructions for both models. There is no appreciable qualitative difference between the pre-patterns.}
    \label{fig:robustness}
    \vspace{-0.5cm}
\end{figure*}

\subsection{Offloading information between pre-patterns and development}
% \subsection{Expressivity of the pre-pattern function} -> wrong title

An interesting question is how information is distributed between pre-patterning and development (note that, in principle a powerful enough pre-pattern function can generate the final stimuli on its own). To analyse this, we measure how information is traded off between the two components for different levels of complexity in the SIREN function using three measures: the normalized mutual information between a (pre-) pattern and its target, pixel-wise linear $R^2$ and structural similarity index measure (SSIM, \cite{wang_image_2004}).

The plots in Figure~\ref{fig:analysis-info-trade} show that indeed the final pattern is not easily predictable from the initial pre-pattern but they are from the final reconstruction. On the other hand, a more surprising result is the fact that the information content of pre-patterns generated by more powerful networks does not substantially increase, even though the final pattern for these are closer to the target. This finding is replicated in the SSIM, which even shows a degradation of the structural similarity between pre-pattern and target. Taken together we interpret these findings as suggesting that pre-patterns do not necessarily make the initial developmental state be closer to the target. Instead they learn to set the NCA on a path that is easier to traverse in developmental space.

\subsection{Robustness to cellular noise}
% \subsection{Pre-patterns indirectly lead to more robustness to noise}

One major issue that any distributed system must overcome is the unreliability of its components, i.e., \textit{Does using a pre-pattern offer any advantages when dealing with this issue?} To explore this we introduce unreliability in the NCA architecture as in \citep{mordvintsev_growing_2020}, where units only update their states with a given probability. In this case, we set the reliability during training to $0.75$ and train three different seeds of the model. At test time, we vary the update probability values in $[0.25, 0.75]$ range at intervals of $0.05$. Thus, the model must deal with increased unreliability as we diminish the update probability. See results in Figure~\ref{fig:robustness}, left.

We observe that for the training condition (in blue), the model achieves a low error rate, consistent with the reconstructions that we showed before. When we vary the update probability at test time, the model's performance shows very little degradation even for low levels of reliability. 

But how good is this relative to an alternative model? To test this, we use a control model which can also generate multiple targets but without the benefit of being initialised using a pre-pattern. Specifically, we use the goal-directed NCA (GoalNCA) from \cite{sudhakaran_goal-guided_2022}. Here, generating multiple patterns is achieved by conditioning the update function using the embedding from the input. Since the GoalNCA does not have the benefit of having access to the pre-pattern to break symmetry, it is afforded additional steps to converge. Specifically, for $64\times64$ targets the model is afforded between $32+16=48$ and $32+16+16=64$, where the $32$ comes from the number of steps required to propagate information from the boundaries to the centre of the substrate and the extra $16$ to $32$ steps are the number we give to the NCA when SIREN pre-patterns serve as its initial conditions.

The results for the control model make it clear that having access to the pre-pattern helps the model cope with the increased level of noise even though it has less computational budget available. The reason is that unreliability during training forces the pre-patterns to initialise each cell along a path that requires less information to be propagated via self-organisation. On the other hand, GoalNCA has no other option but to propagate information about boundary conditions via self-organisation. When cell communication becomes unreliable, the whole process breaks down.

Indeed we can see that this is the case if both models use a deterministic update (that is, update probability equal to $1.0$). The results in Figure~\ref{fig:robustness}, panel b, show that in this case performance degrades significantly for both models, even if the model with SIREN pre-patterns is less affected by it (though the difference is not significant).

One plausible hypothesis is that pre-patterns are qualitatively different between the two variants of the NCA with SIREN pre-patterns when trained with different update probabilities. However, we found no evidence of this. Indeed the reconstructions in panels b) and c) show that across both conditions there is no discernible difference between the two sets of pre-patterns. This suggests that instead what happens is that the NCA learn to use this information in a qualitatively different way that allows it to overcome the unreliability in cell-to-cell communication. This is consistent with prior work showing that NCA trained with asynchronous (stochastic) updates learn more well-behaved dynamics that remain stable under changes to the cell update regime, whereas their deterministically-updated counterparts do not \citep{Niklasson2021Jul}.

\subsection{Pre-patterns increase model storage capacity}

\begin{figure}[t]
    \centering
    \includegraphics[width=\linewidth]{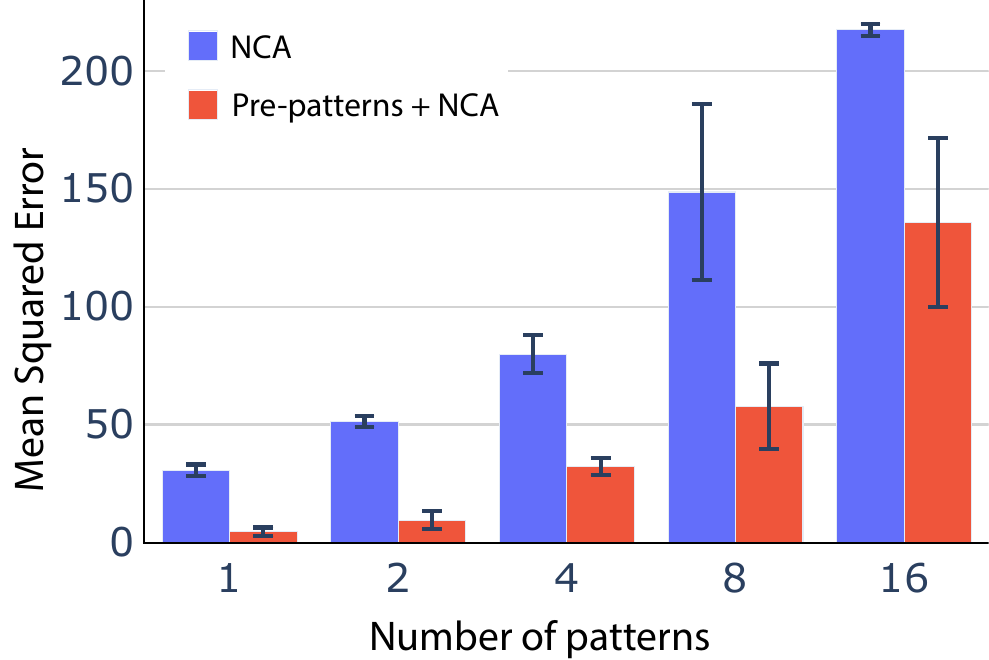}
    \caption{\textbf{Analysing model capacity}. We compare reconstruction error when using pre-patterns generated with SIREN against GoalNCA --- a purely self-organising control --- as both are tasked with encoding an increasing number of patterns ($1$ to $16$ in powers of two). At comparable total parameter budgets, the SIREN+NCA model is consistently able to store more patterns than GoalNCA, using reconstruction MSE as a proxy. This is the memory-compute trade-off in action: parameters that would otherwise have to encode propagation rules powerful enough to disambiguate every target (compute) are spared by storing per-target positional information in the initial state (memory), so the same budget encodes more patterns. Each condition is run with three different random seeds.}
    \label{fig:capacity}
    \vspace{-0.3cm}
\end{figure}

One of the main benefits of compositionality is that it allows for factorized representations which reuses components. Such a system is thus capable of encoding or representing a much larger set of patterns than one that isn't compositional. As we argued before, a system that pairs an NCA with SIREN pre-patterns as initial conditions can be seen from this perspective as well. Instead of composing parts of a whole, it composes \emph{functions}, each of which has a separate capability. So far we have shown that this approach has benefits in terms of robustness and, moreover, both components seem to encode separate information content. 

This leads us to the question of whether this combination of pre-pattern and self-organisation also exhibits some of the hallmarks of compositionality. First we start by considering the problem of the system's capacity. We can measure it by how many patterns can be encoded in a system. To test this there are two issues to consider: the size of the model and how to measure the number of parameters stored. For the first issue we need a model that is sufficiently powerful to encode at least one input, but which shows appreciable degradation as we increase the number of patterns. 

\begin{figure*}
    \centering
    \includegraphics[width=\linewidth]{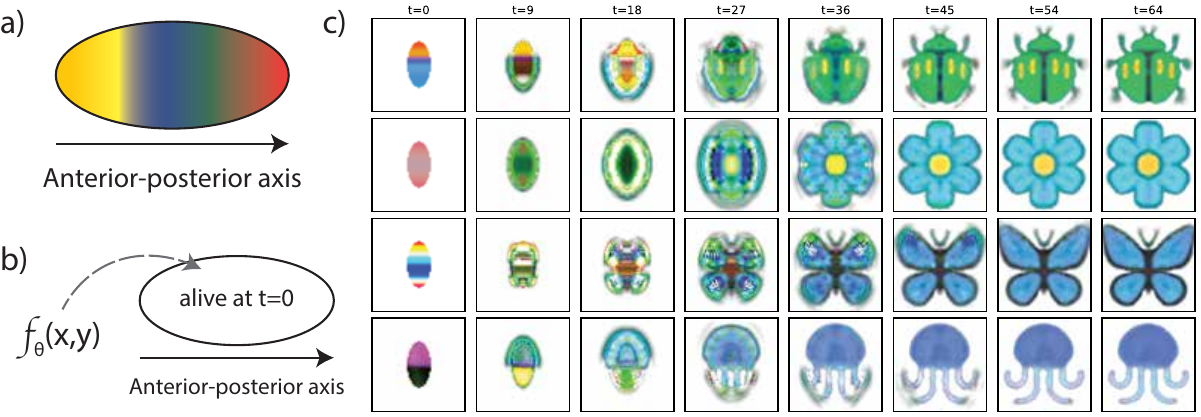}
    \caption{\textbf{Capturing the structure of biological pre-patterns}. \textbf{a)} A schematic of differential expression of homeobox genes during early development as a consequence of maternal-effect genes, based on \citep{schroeder_transcriptional_2004}. \textbf{b)} How a pre-pattern function (in this case SIREN) can generate the pre-patterns. \textbf{c)} The trained model showing 4 different examples where the pre-pattern exhibits the same kind of sinusoidal structure. We can observe how they break symmetry and enable the propagation of developmental signals across the substrate.}
    \label{fig:growth}
    \vspace{-0.5cm}
\end{figure*}

To this end we reduce the model's representational capacity substantially by reducing the pre-pattern MLP width to $16$, the embedding dimensionality to $4$ and the NCA's update function MLP width to $32$ (see Figure~\ref{fig:model-def}). For the control we perform the same reduction where appropriate. For the second issue we would ideally follow a similar approach as when measuring the capacity of models such as Hopfield Networks \cite{hopfield_neural_1982}. However, the continuous nature of the image patterns that we are using as testing means that counting correct bits is not possible. Thus we resort once again to measuring the mean-squared error between the recovered patterns and the targets.

To test how models behave as their representation capacity is tested, we train them on five different conditions with $1$, $2$, $4$, $8$ and $16$ patterns. We also select patterns that are not too similar i.e. only one of the blossoms and one of the butterflies. The results shown in Figure~\ref{fig:capacity} clearly illustrate the advantage of using a pre-pattern in terms of encoding capacity; both models degrade in performance as the number of patterns increases, as expected, but the degradation is much less pronounced when pre-patterns serve as initial conditions. This happens even though the model has a similar number of parameters to the control ($\approx1400$ vs $\approx1500$).

\section{Pre-patterns in growing systems}
% \subsection{Considering the problem of growth}

So far we have considered properties of a model that performs self-organisation over a fixed developmental substrate: the number of units and the area that they occupy is fixed throughout the developmental or generative process. However growth is a fundamental characteristic of biological development (and incidentally is a main source of motivation for this work) which we have so far not considered in this context. Such an omission obeys a practical reason: it is much easier to analyse and understand a system by reducing it to its essential components. Having highlighted the benefits of systems that use pre-patterns to offload computation throughout the previous sections, here we introduce growth into the system.

To introduce growth into the system we define an initial substrate from which the system must expand in order to grow the final morphology. Taking inspiration again from biological development we define such starting substrate as an ellipse (which resembles an egg, see Figure~\ref{fig:growth} panel a). The pre-pattern function can generate values within this substrate while the rest of the potential cells are \textit{turned off} (see Figure~\ref{fig:growth} panel c). The rest of the system works in much the same way as the growing NCA \citep{mordvintsev_growing_2020}. Notice however that in those cases the starting condition is a single cell. Indeed, to break symmetry when an isotropic NCA is considered, subsequent work had to rely on multiple cells with distinct identities. Such an approach resembles our own focus on pre-patterns but instead highlighted the capacity of self-organisation with very little information — though at the expense of robustness and stability. Modelling cells as isotropic units is however desirable as this is the mechanism most often found in biological development.

Simulating the system after training we observe results as the one shown in panel c of Figure~\ref{fig:growth}. One surprising fact is that across all runs the pre-patterns exhibited the same type of qualitatively similar sinusoidal-like shape. In principle there is no reason for this to be the case since the target morphologies do not possess such a shape. Even more strikingly, this is very similar to spatial arrangement of gene expression for homeobox genes during early development. This suggests that such sinusoidal patterns are likely optimal for symmetry breaking when pre-patterns follow gradients at different rates and future growth is involved (since we did not observe this effect in our previous simulations).

We can see how in this case both mechanisms, pre-patterning and self-organisation interplay to create the final morphology: the initial pre-pattern establishes ways to break symmetry (clearly seen in the example of the jellyfish at the end), but given the complexity of the target shapes signals are regularly propagated to coordinate development downstream. We conclude that such systems are thus promising avenues for modelling biological processes of early development \citep{schroeder_transcriptional_2004, bender_molecular_1983}.

\section{Discussion}

In this work we have set out to study how information is distributed between spatially structured initial conditions and developmental dynamics implemented using self-organising processes. Across a series of computational simulations we have shown that while in principle self-organisation on its own can achieve similar developmental tasks, it requires more computational resources and limits the robustness and expressivity of the system. 

Counter-intuitively we find that pre-patterns do not work by necessarily setting the system closer to its target but by biasing the trajectory of development. This is a critical distinction which is not obvious: Figure~\ref{fig:model-def}b suggests that with greater expressivity the pre-pattern learns initialisations that disambiguate development when jointly trained with the NCA, even though it could just reproduce the target directly. We hypothesize that this is a consequence of using the same dynamics to generate multiple patterns. Where the system lands on this trade-off depends on both the pre-pattern generator's expressivity, which we vary here, and the optimisation setup, which we hold fixed; the latter would be expected to yield solutions with a different balance of memory and compute. It would be interesting to ask how this is manifested in more realistic systems where differences are not only in the initial conditions, but in the patterns of dynamics themselves (i.e. due to different gene-regulatory networks).

Indeed as shown in Figure~\ref{fig:analysis-info-trade}, increasing expressivity of the pre-pattern did not lead to significant transfer of information to the pre-pattern function. Furthermore, when reducing the reliability of cells in the system, the patterns do not look qualitatively different from the ones with perfectly reliable cells (Figure~\ref{fig:robustness}). One plausible explanation is that the developmental component learns to better exploit locality and the information already provided by the pre-pattern. Finally, the capacity experiments show that this composition of functions between the pre-patterning and self-organisation unlocks more efficient use of parameters in the system (Figure~\ref{fig:capacity}). This makes the memory-compute trade-off concrete: information transferred into the initial state (memory) is information the self-organising process (compute) no longer has to generate from scratch.

Seen through this lens, the three gains we measure are different faces of the same trade-off: robustness improves because information read from the initial state is not corrupted by unreliable propagation; encoding capacity improves because shared positional information need not be re-derived per target; and symmetry breaking is provided up-front rather than at every step of self-organisation. A purely self-organising alternative pays the compute side of this trade-off in full, which is a clear motivation for scaffolding development instead of relying on self-organisation alone to generate complex patterns. While we have mainly focused on biological development, we believe that these insights hold across many other modalities where self-organisation needs to be constrained in order to scale to generate more complex structures.

In the context of biological development, our results provide, to our knowledge, the first computational model capable of jointly learning top-down pre-patterns alongside a shared self-organising developmental program. Contrary to most biological studies, which focus on describing \emph{how} pre-patterning occurs, we directly compare against an alternative that cannot exploit scaffolding --- yielding an information-theoretic account for \emph{why} such pre-patterns are prevalent in complex developmental systems, unifying insights from positional information \citep{wolpert_positional_2016} and self-organisation \citep{gershenson_self-organizing_2025}.
% Returning to the issue of biological development, our results offer a computational account of why morphogenetic gradients are fundamental for development \cite{meinhardt_models_2008}. However, contrary to most studies in biology which focus on describing how this process happens, we directly compare to an alternative model which cannot rely on scaffoldings. Thus we offer a computational account for why such pre-patterns are prevalent in complex systems unfying insights from positional information \citep{wolpert_positional_2016} and self-organistion \citep{gershenson_self-organizing_2025}.

Of especial interest are the simulations where growth is involved (Figure~\ref{fig:growth}). It is not obvious that when optimized to generate patterns the system would use sinusoids that extend along the anterior posterior axis of development as in models of drosophila \citep{bender_molecular_1983}, yet it does. This result suggests that such an organisation is likely optimal or near optimal, which is why it was selected via evolution.

\subsection{Limitations}

There are many processes that intervene in biological development which we are not considering. Perhaps more critically, we are optimizing a system to fit particular pre-patterns. This allows us to make an optimality argument in favour of our conclusions but biology is not necessarily optimal. One alternative would have been to use evolutionary algorithms instead of back-propagation. However, we argue that current evolutionary algorithms don't exactly capture the nuances involved with real evolution and moreover make it substantially harder to fit the system. Future work must explore plausible mechanisms for co-evolving both the pre-patterns and the self-organising system.

Related to this issue of using back-propagation, we hypothesized that using scaffoldings to guide self-organisation would lead to increased evolvability, making similar patterns more accessible to the system. However, when testing this the results were underwhelming. We argue that this is more an issue with back-propagation itself than with the core principle. Indeed, Deep Learning approaches tend to rely on massive amounts of data in order to capture patterns that enable generalisation to emerge. Scaling the model is also made difficult by the necessity to propagate information through time. Future work must address these issues in order to enable the study of more complex processes.

\subsection{Related work}

Using pre-generated patterns to scaffold future development has previously been studied in the context of Neuroevolution. Perhaps the most popular example is the use of HyperNEAT as a way to create pattern that determines where units from a neural network should be laid out \citep{risi_evolving_2010}. Indeed, SIRENs and ES-HyperNEAT can be both be seen as examples of implicit representations since their inputs are coordinates in a substrate. Furthermore, ES-HyperNEAT has been used to find the connectivity for a biologically plausible early vision system \citep{risi_guided_2014}. The system that we use here is similar in spirit, except that instead of generating an information processing network we are developing the morphologies represented as spatial patterns. 

On the other hand, implicit representations have not been used in this manner in the context of self-organising systems. \cite{pajouheshgar_noisenca_2024} proposed the use of implicit representations as a way to scale up NCA pattern generation by interpolation between self-organizing nodes, but this is fundamentally different in spirit to its use in the current work.

% A different way in which the idea of pre-patterning can be seen is that of meta-learning. Systems such as MAML \citep{finn_model-agnostic_nodate} in reinforcement learning work in a similar way: learn a good set of initial conditions that allow a system to be fine-tuned into reaching many different targets. 
% Similarly, our work is similar to that which tackles the problem of discovering governing equations using Deep Learning approaches, which has become a popular method for modelling real phenomena which physics constraints \cite{raissi_physics-informed_2019}. However, such approaches avoid learning initial conditions since they are usually fit to one dynamical process at a time, not several as in our case (but see \citet{cox_initial_1995} and \citet{mccabe_learning_nodate} for examples).

Perhaps the work that is closest to our own is \citep{hartl_evolutionary_2024}, which also learns a pre-pattern along with a developmental system via evolution. However, in this case the system only needs to target one pattern (a flag). Our work thus extends this into studying how common dynamics can be guided towards different morphological targets which offers a different perspective of shared developmental processes.

\subsection{Future work}

Beyond the technical issues related to optimisation in complex models mentioned above, future work should explore how pre-patterns can unlock greater evolvability in self-organising systems. Indeed we have shown the benefits of composing functions, but not how the composition can happen in the structure of the initial conditions. Studies have shown that it is possible to obtain differentiated modules without explicit modularity in the generation process \citep{khona_global_2025}, so in theory this should be possible with the current approach.

While we have focused on a two step procedure where pre-patterns are laid and self-organisation proceeds afterwards, these processes likely alternate during development as the system descends down the developmental pathway. Thus an interesting future question is how to model this hierarchical interplay between scaffolding and self-organisation and how they can lead to increased complexity by alternating between the two.

\section{Acknowledgements}

The authors would like to thank the members of the REAL and Uni-Bio labs, especially Eleni Nisioti, Ala Trusina and Teresa Knudsen for helpful comments and suggestions.

MM is funded by the European Union (ERC, GROW-AI, 101045094), EN and JS are funded by the Novo Nordisk Foundation Synergy Grant \textit{REPROGRAM} number NNF23OC0086722.

% \pagebreak
\footnotesize
\bibliographystyle{plainnat}
\bibliography{mncas} % replace by the name of your .bib file

\end{document}